# Sign Language Detection

Amey Chavan, Shubham Deshmukh, Favin Fernandes

Department of Electronics and Telecommunication, Vishwakarma Institute of Technology, Pune

amey.chavan18@vit.edu, shubham.deshmukh18@vit.edu, favin.fernandes18@vit.edu

**Abstract:** With the advancements in Computer vision techniques the need to classify images based on its features have become a huge task and necessity. In this project we proposed 2 models i.e. feature extraction and classification using ORB and SVM and the second is using CNN architecture. The end result of the project is to understand the concept behind feature extraction and image classification. The trained CNN model will also be used to convert it to tflite format for Android Development.

## I. Introduction

American Sign Language (ASL) is a complete, natural language that has the same linguistic properties as spoken languages, with grammar that differs from English. ASL is expressed by movements of the hands and face. It is the primary language of many North Americans who are deaf and hard of hearing, and is used by many hearing people as well.

There is no universal sign language. Different sign languages are used in different countries or regions. For example, British Sign Language (BSL) is a different language from ASL, and Americans who know ASL may not understand BSL. Some countries adopt features of ASL in their sign languages.

No person or committee invented ASL. The exact beginnings of ASL are not clear, but some suggest that it arose more than 200 years ago from the intermixing of local sign languages and French Sign Language (LSF, or Langue des Signes Française). Today's ASL includes some elements of LSF plus the original local sign languages; over time, these have melded and changed into a rich, complex, and mature language. Modern ASL and modern LSF are distinct languages. While they still contain some similar signs, they can no longer be understood by each other's users.

ASL is a language completely separate and distinct from English. It contains all the fundamental features of language, with its own rules for pronunciation, word formation, and word order. While every language has ways of signaling different functions, such as asking a question rather than making a statement, languages differ in how this is done. For example, English speakers may ask a question by raising the pitch of their voices and by adjusting word order; ASL users ask a question by raising their eyebrows, widening their eyes, and tilting their bodies forward.

Just as with other languages, specific ways of expressing ideas in ASL vary as much as ASL users themselves. In addition to individual differences in expression, ASL has regional accents and dialects; just as certain English words are spoken differently in different parts of the country, ASL has regional variations in the rhythm of signing, pronunciation, slang, and signs used. Other sociological factors, including age and gender, can affect ASL usage and contribute to its variety, just as with spoken languages.

Fingerspelling is part of ASL and is used to spell out English words. In the finger spelled alphabet, each letter corresponds to a distinct handshape. Fingerspelling is often used for proper names or to indicate the English word for something.

Parents are often the source of a child's early acquisition of language, but for children who are deaf, additional people may be models for language acquisition. A deaf child born to parents who are deaf and who already use ASL will begin to acquire ASL as naturally as a hearing child picks up spoken language from hearing parents. However, for a deaf child with hearing parents who have no prior experience with ASL, language may be acquired differently. In fact, 9 out of 10 children who are born deaf are born to parents who hear. Some hearing parents choose to introduce sign language to their deaf children. Hearing parents who choose to have their child learn sign language often learn it along with their child. Children who are deaf and have hearing parents often learn sign language through deaf peers and become fluent.

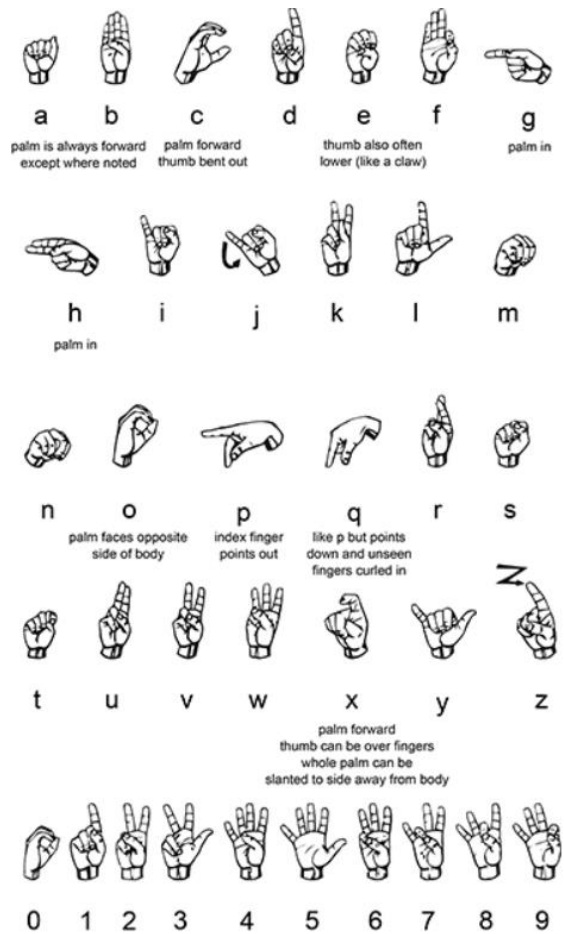

In order to bridge the Gap between these children and adults the use of Computer vision and Feature Extraction techniques are important as there is no necessity of a 3rd Party to help with the Translation of the ASL Language.

Therefore, to build a system that can recognize sign language will help the deaf and hard-of-hearing better communicate using modern-day technologies. In this paper, we will go through architectures of CNN and ORB to see how it performs on classifying the Sign Language.

## II. Literature Review

In this paper [1] A real-time sign language translator is an important milestone in facilitating communication between the deaf community and the general public. We hereby present the development and implementation of an American Sign Language (ASL) fingerspelling translator based on skin segmentation and machine learning algorithms. We present an automatic human skin segmentation algorithm based on color information. The YCbCr color space is employed because it is typically used in video coding and provides an effective use of chrominance information for modeling the human skin color. We model the skin-color distribution as a bivariate normal distribution in the CbCr plane. The performance of the algorithm is illustrated by simulations carried out on images depicting people of different ethnicity. Then Convolutional Neural Network (CNN) is used to extract features from the images and Deep Learning Method is used to train a classifier to recognize Sign Language.

In this paper [2] the authors introduce a Sign Language recognition using American Sign Language. In this study, the user must be able to capture images of the hand gesture using web camera and the system shall predict and display the name of the captured image. They used HSV color algorithm to detect the hand gesture and set the background to black. The images undergo a series of processing steps which include various Computer vision techniques such as the conversion to grayscale, dilation and mask operation. The features extracted are the binary pixels of the images. They make use of Convolutional Neural Network(CNN) for training and to classify the images. They were able to recognize 10 American Sign gesture alphabets with high accuracy. The model has achieved a remarkable accuracy of above 90%

In this paper [3], the hand regions segmented from the depth image using the Microsoft Kinect Sensor in the cluttered environment. The depth image obtained is then used to implement supervised machine learning by extracting and training the features of images. Here, by comparing various methods, it is depicted that ORB (Oriented FAST and Rotated BRIEF) outruns others in terms of accuracy. ORB is invariant to scale, rotation, and lighting conditions. ORB is also fused with various classification techniques to gain the optimum result. The method is applied to images of ISL 0–9 and is also compared with some standard datasets. Tuning of the ORB with k-NN classification produces an average recognition accuracy of 93.26% with ISL dataset.

### III.Methodolgy

**1)Dataset**

For this project, the data was collected from Kaggle having 36 classes comprising of A-Z alphabets in lower case and 0-9 numbers (50 images each). The images for CNN architecture was resized to 224x224 and for ORB the images were resized to 512x512.

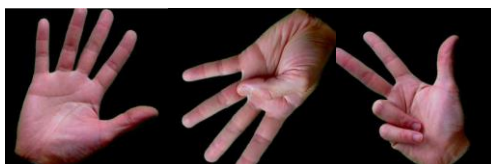

fig1. American Signs of 5,4 and 3

**2)Method**

**A) Feature Extraction and Classification using ORB and Decision Tree:**

ORB:

This algorithm was brought up by Ethan Rublee, Vincent Rabaud, Kurt Konolige and Gary R. Bradski in their paper ORB: An efficient alternative to SIFT or SURF in 2011. As the title says, it is a good alternative to SIFT and SURF in computation cost, matching performance and mainly the patents.

ORB is basically a fusion of FAST keypoint detector and BRIEF descriptor with many modifications to enhance the performance. First it uses FAST to find keypoints, then apply Harris corner measure to find top N points among them. It also uses pyramid to produce multiscale-features. But one problem is that, FAST doesn't compute the orientation. So what about rotation invariance? Authors came up with following modification.

It computes the intensity weighted centroid of the patch with located corner at center. The direction of the vector from this corner point to centroid gives the orientation. To improve the rotation invariance, moments are computed with x and y which should be in a circular region of radius r, where r is the size of the patch.

Now for descriptors, ORB use BRIEF descriptors. But we have already seen that BRIEF performs poorly with rotation. So what ORB does is to "steer" BRIEF according to the orientation of keypoints. For any feature set of n binary tests at location $(x_i, y_i)$, define a $2 \times n$ matrix, S which contains the coordinates of these pixels. Then using the orientation of patch, θ, its rotation matrix is found and rotates the S to get steered(rotated) version $S_\theta$.

ORB discretize the angle to increments of $2\pi/30$ (12 degrees), and construct a lookup table of precomputed BRIEF patterns. As long as the keypoint orientation θ is consistent across views, the correct set of points $S_\theta$ will be used to compute its descriptor.

BRIEF has an important property that each bit feature has a large variance and a mean near 0.5. But once it is oriented along keypoint direction, it loses this property and become more distributed. High variance makes a feature more discriminative, since it responds differentially to inputs. Another desirable property is to have the tests uncorrelated, since then each test will contribute to the result. To resolve all these, ORB runs a greedy search among all possible binary tests to find the ones that have both high variance and means close to 0.5, as well as being uncorrelated. The result is called rbrief.

For descriptor matching, multi-probe LSH which improves on the traditional LSH, is used. The paper says ORB is much faster than SURF and SIFT and ORB descriptor works better than SURF. ORB is a good choice in low-power devices for panorama stitching etc.

Kmeans for dimensionality Reduction:

K-Means Clustering is an Unsupervised Learning algorithm, which groups the unlabeled dataset into different clusters. Here K defines the number of pre-defined clusters that need to be created in the process, as if K=2, there will be two clusters, and for K=3, there will be three clusters, and so on.

It allows us to cluster the data into different groups and a convenient way to discover the categories of groups in the unlabeled dataset on its own without the need for any training.

It is a centroid-based algorithm, where each cluster is associated with a centroid. The main aim of this algorithm is to minimize the sum of distances between the data point and their corresponding clusters.

The algorithm takes the unlabeled dataset as input, divides the dataset into k-number of clusters, and repeats the process until it does not find the best clusters.

This Algorithm Decreases the dimension of a data by clustering the data points into no of clusters given by the user.

Decision Tree Classifier:

Decision Tree is a Supervised learning technique that can be used for both classification and Regression problems, but mostly it is preferred for solving Classification problems. It is a tree-structured classifier, where internal nodes represent the features of a dataset, branches represent the decision rules and each leaf node represents the outcome.

In a Decision tree, there are two nodes, which are the Decision Node and Leaf Node. Decision nodes are used to make any decision and have multiple branches, whereas Leaf nodes are the output of those decisions and do not contain any further branches.

The decisions or the test are performed on the basis of features of the given dataset.

It is called a decision tree because, similar to a tree, it starts with the root node, which expands on further branches and constructs a tree-like structure.

In order to build a tree, we use the CART algorithm, which stands for Classification and Regression Tree algorithm.

A decision tree simply asks a question, and based on the answer (Yes/No), it further split the tree into subtrees.

Below diagram explains the general structure of a decision tree:

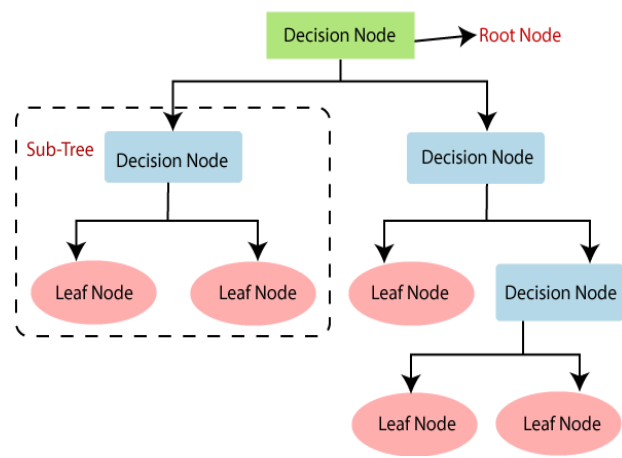

Proposed method:

Algorithm

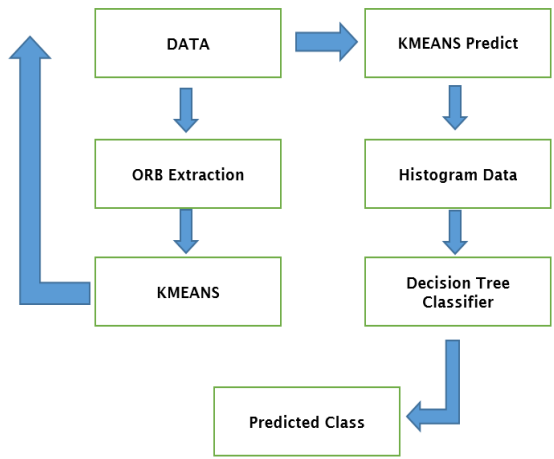

Working:

The images are first resized and converted to gray for uniformity and the feature extraction is to be done on gray images only. The feature extracted are the ORB features, these features of the entire class in stored in a feature vector which is then saved in a csv file. The process runs for all the classes.

After Feature Extraction is done, All the feature vectors of the classes are appended and using kmeans, 5 clusters are fit and the model is saved.

The saved kmeans is then used to predict the kmeans value for each image of a class, this creates 4 feature vectors Columns=5.

After the generation of kmeans predicted feature for each class, the labels are assigned to each row i.e. 0,1,2,3 etc. for all the classes.

This feature vectors are appended and fed to the decision tree classifier, the max depth of the DT classifier is predicted by running the classifier until the accuracy of the model is highest.

**B) CNN Architecture:**

A Convolutional Neural Network or CNN is a special type of neural networks that specializes in processing data that has a grid-like topology, such as an image. Neural Network is well known for finding out patterns in various kind of datasets, convolutional is particularly used for images as image being huge chunk of data convolution operation can extract meaning information from it without losing much information and perform task quickly. Any CNN model is comprised of three basic layers: convolution layer, pooling layer, fully connected layer. In convolutional layer it performs a dot product between two matrices, where one matrix is the set of learnable parameters otherwise known as a kernel, and the other matrix is the restricted portion of the receptive field. The kernel is used for feature extraction for finding out various features in an image such as vertical, horizontal edges etc., these set of kernels are convolved with the input image. The kernel is spatially smaller than an image but is more in-depth. This means that, if the image is composed of three (RGB) channels, the kernel height and width will be spatially small, but the depth extends up to all three channels. Pooiling layers are used to extract most useful information from the convolved outputs, various other commands are used to prevent overfitting. Fully connected layers are used perform normal neural network operations to detect patterns from the outputs of previous layers. Along with convolution layers' various activation functions like relu, sigmoid, and softmax are used to formulate conv layers output. The proposed CNN architecture used for sign language detection uses 4 Conv2d ,4 MaxPool2d, 4 Dropout, 1 GlobalAveragePooling2d, and 3 Flatten Dense layers finally giving us 36 multi class classification output using softmax. Using Adam optimizer as an optimization technique along with categorical crossentropy as an accuracy metrics the model is fitted and achieving training accuracy as 96% and testing accuracy as 92%. Now the weights and biases trained through the proposed architecture are saved in keras model format (.h5 model ) and later can again be converted to tflite model from which it has been used to detect sign language through mobile android application in real time.

### C) Android Application

The trained CNN model is then saved in the .H5 model and converted to tflite format. The. tflite format supported by Tensorflow can be used to classify images using Android Studio.

### IV. Results

1) ORB:

Feature vector:

Reduced Dimension:

Decision Tree Classifier Accuracy: 20%

2) CNN

Model Summary:

Training Accuracy for each Epoch:

Model Accuracy:

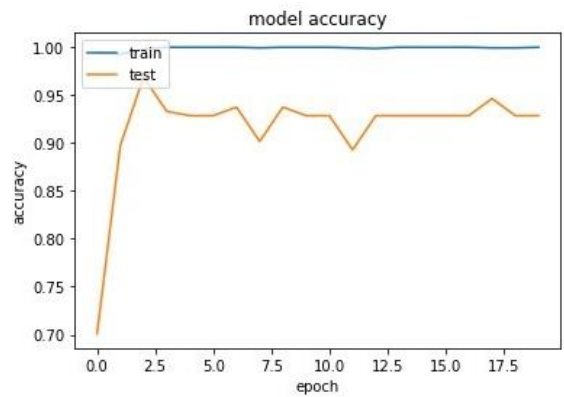

Model loss:

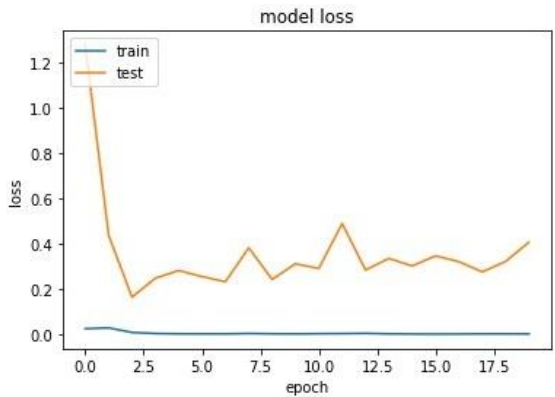

3)Android Application

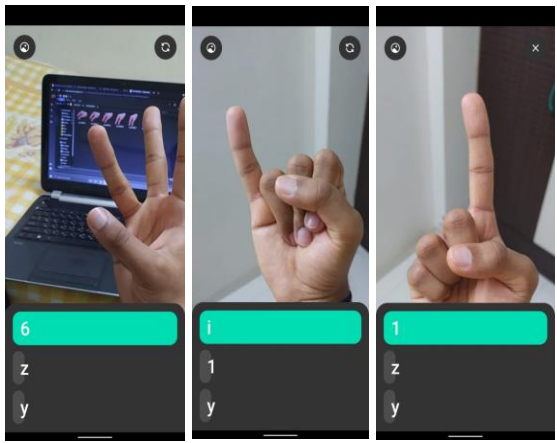

## V.Conclusion

ORB feature extraction is computationally faster for feature extraction but the accuracy of the model for classification is very low. CNN on the other hand has provided good results in terms of ORB. The CNN model can also be converted to many other formats in order to implement trained models by users in Web, Android and Flutter programs.
At the very least CNN techniques has a upper hand when it comes to feature extraction and classification together.

## Refernces: